# Hey ASR System! Why Aren't You More Inclusive?
## Automatic Speech Recognition Systems' Bias and Proposed Bias Mitigation Techniques. A Literature Review.

Mikel K. Ngueajio [1[0000-0002-9696-2372]] and Gloria Washington [2[0000-0002-8524-3559]]

Department of Computer Science, Howard University, Washington DC 20059, USA
[1]mikel.ngueajio@bison.howard.edu, [2]gloria.washington@howard.edu

**Abstract.** Speech is the fundamental means of communication between humans. The advent of AI and sophisticated speech technologies have led to the rapid proliferation of human to computer-based interactions, fueled primarily by Automatic Speech Recognition (ASR) systems. ASR systems normally take human speech in the form of audio and convert it into words, but for some users it cannot decode the speech and any outputted text is filled with errors that are incomprehensible to the human reader. These systems do not work equally for everyone and actually hinders the productivity of some users. In this paper, we present research that addresses ASR biases against gender, race, and the sick and disabled, while exploring studies that propose ASR debiasing techniques for mitigating these discriminations. We also discuss techniques for designing a more accessible and inclusive ASR technology. For each approach surveyed, we also provide a summary of the investigation and methods applied, the ASR systems and corpora used, the research findings, and highlight their strengths and/or weaknesses. Finally, we propose future opportunities for Natural Language Processing researchers to explore in the next level creation of ASR technologies.

## 1 Introduction

Just a few decades ago, the thought of holding a meaningful conversation with a device felt so far-fetched. Thanks to the advent of innovative AI technology for speech and language processing, called Automatic Speech Recognition, computers are now capable of listening and understanding context and nuances in human language. Smart Voice Assistants (SVA) are so readily available to the general public because of their seamless integration into a wide variety of omnipresent hardware devices like smartphones, PCs, in-car-voice-command and smart home systems, and software e.g., web searches, automatic subtitling, Interactive Voice Response systems. The number of SVA in use has been estimated at 8 billion and is predicted to triple in the next few years [1]. These stats are even more feasible owing to the global lockdown as a result of the novel COVID-19 pandemic outbreak and its afflictions on traditional interpersonal interaction [2-4]. With more people subscribing to new forms of communication, it is crucial to understand the roles and implications of these devices' use for individuals and society. As technology evolves and brings about new waves of flashy devices intended to make "lives better and easier," research has shown that these expected outcomes are not always experienced, in the same capacity by everyone [5-9].

This work provides a comprehensive review of bias in ASR systems via the following contributions: We (1) Summarize recent studies of ASR systems exhibiting, promoting, and/or exacerbating cultural, societal, and health bias and (2) Review proposed approaches to help mitigate biases in these systems. The survey was conducted with the aims of collecting and highlighting the greatest number of research and proposed approaches in this area. The relevant study selection process involved 1) Searching the internet for specific keyword analyzing "Bias or discrimination" in ASR, NLP, Voice, language, and Speech technology, referring, or related to the topic area, 2) Selecting and gathering studies from different databases such as Google Scholar, ACM Digital Library, Scotus, ArXiv, emphasizing mainly on studies from Computer Science, Engineering and/or Psychology fields,  and 3) Manually inspecting and reading selected papers, focusing primarily on research conducted in the past 5 years, prior to March 2022, with the aims of categorizing them as Racial, gender or Disability ASRs bias.

Recent survey papers addressing AI systems' bias or debiasing techniques focus mainly on tackling this issue in Natural Language Processing (NLP) and /or emphasize on unique genre of discrimination. For instance, Blodgett et. al [10] provide a critical survey of 146 research focusing exclusively on Bias in NLP; Sun et. al [11] provide a comprehensive literature review of research addressing and proposing methods for mitigating gender bias in NLP while Garrido-Munoz et. al [12] address recent development on Bias in Deep NLP.  To the best of our knowledge, there has been no other survey research that addresses ASR systems' Bias exclusively, and from the various spectrum namely race, gender, sick and disability, as presented in this paper.

The rest of the paper is structured as follows: - the next section provides the literature review of research surrounding biased and discriminatory ASR systems. Next, we provide a survey of proposed approaches for debiasing ASR systems and finally we provide conclusions and discuss opportunities for next steps in section 5.

## 2    BIASES IN ASR SYSTEMS

This section reviews research that has studied ASR systems and reported patterns of discrimination against race, gender, physically ill, and People with Disabilities (PWD).

### 2.1    ASR Systems Exhibiting Racial, Social, and Cultural Bias

ASR systems have come a long way since the IBM shoebox of 1961 which could perform mathematical functions and recognize 16 languages. Nowadays, these systems are more sophisticated and ubiquitous but are also, imperfect. Growing bodies of research are revealing them as being biased or as accentuators of societal stereotypes mainly against individuals from marginalized communities. For instance, a study examining 5 ASR transcription systems [13] developed by Amazon, Apple, Google, Microsoft, and IBM, proves that the systems exhibit racial bias. The experiment is based on conversation speeches from 42 white and 73 Black speakers of average age 45yrs collected from the Corpus of Regional African American Language (CORAAL) and the Voice of California (VOC) Datasets. The 5 to 50s long manually transcribed subset of audios used are matched across these datasets based on the speakers' age, gender, and audio duration, hence resulting in 2141 audio snippets, 44% of which are from male speakers. The manual transcription generated is used as ground truth for evaluating the ASR transcription quality using the Word Error Rate (WER) metric. Initially, an average WER for the automated transcription is computed against the matched audio snippets of white and Black speakers, and preliminary results show an average WER almost doubled for Black speakers, with a significant WER disparity between Black males and females, contrasted by comparable WER for white genders. To determine if speakers' locations contribute to the accuracy gap, a comparative study of average WER for Black speakers from Black neighborhoods (Princeville, Washington DC, and Rochester) with that of white speakers from the predominantly white neighborhoods of Sacramento and Humboldt County, is analyzed. This investigation reveals a wider median error rate for Black neighborhoods. An inquiry into the density of African American Vernacular English (AAVE) in these Black neighborhoods is also analyzed to help pinpoint the root cause of this disparity and the findings reveal a positive correlation between AAVE language density and speakers' WER per neighborhood. From another perspective, a detailed exploration of the language and acoustic models underlying the ASR systems is also undertaken, by creating an aggregate collection of unique words from all speakers for each ASR transcription system and then comparing the percentage of words, absent from the ASRs' vocabulary per race. Greater proportions of utterances from Black speakers could be found in the machines' vocabulary, helped rule out the lack of proper language grammar in Black speakers' utterances. However, an investigation into the ASR systems' acoustic models revealed that WER for Black speakers was twice those of whites' even for identical utterances. With no knowledge of the exact language models used to power the commercial ASR systems considered in this study, the authors evaluate the racial disparity of 3 State-Of-The-Art (SOTA) Language Models (LM) namely, Transformer-XL, GPT, and GPT-2, and found an overall lower average perplexity i.e., better performance for Black speakers despite the LM showing statistical preferences for standard English. The author believes the discrepancies reports in the result could be a product of regional linguistic variation and the study does not say, if the ASR systems investigated matched those used by the companies' virtual assistant devices. Martin and Tang [14] attempt to spot the underlying causes of racial disparity in ASR systems by focusing on the morphosyntactic features - the habitual "Be" of AAVE. In this assessment, 30 instances of the word "Be" are selected from audio snippets of Black AAVE speakers from the CORAAL dataset. Audio clips comprising both the utterance and the speaker's turn in which it occurred are then extracted from each audio, then hand-tagged and grouped as either habitual "Be" (376 instances) or non-habitual "Be" classes (2974 instances). The DeepSpeech [15] and Google Cloud Speech ASR systems are used, and their effectiveness at recognizing instances of the habitual "Be" is facilitated through a four-step semi-automatics annotation process devised to help decide correctness. From this, the WER of full utterance and speaker turns is computed using the Wagner-Fischer algorithm. Furthermore, to determine if the ASR systems are capable of understanding context and inferred meaning from speech with habitual "Be", varied amount of context surrounding the occurrence of "Be" within utterances, speech rate, and noise, is analyzed, while their impacts on the systems' performance is assessed. The experiment is conducted in 2 phases. The vulnerability assessment of the habitual "Be" within these systems inferences and then of the words surrounding a habitual "Be", with the most optimal variable(s) determined at each step. In the end, the results show that both systems perform poorly on the tasks of correctly inferring habitual "Be" than non-habitual "Be", with the Google speech systems outperforming the DeepSpeech on both utterances and turn levels. Both systems demonstrate biases against acoustic and morphosyntactic features of the AAVE, meaning that the systems did not perform well on grammatical features specific to AAVE, hence corroborating [13] findings on the system's inaptitude's on AAVE languages. The research also reports higher error susceptibility for the habitual 'Be' and surrounding words compared to non-habitual "Be" and adjacent words, hence validating the limitation of using traditional WER metric for certain morpho-syntactic gerne of words.

Another perspective presented in [16] rather examines the emotional and behavioral toll these systems' biases afflict on their users. The study attempts to bridge the gap between the socio-linguistics and psychology theories, to help understand fairness in ASR systems. Of the 1865 people involved in the prescreening phase of the experiment, 30 African Americans (AA) English speakers, living in Atlanta, Chicago, Houston, Los Angeles, New Orleans, Philadelphia, or Washington D.C participated in this study, which spanned 2 weeks. They were all required to assess their satisfaction with their current devices, report specific instances of dissatisfaction with the devices; record all interaction with the devices daily, and then, complete a set of random pre-assigned tasks, reporting their experience via a 60s video recording and answering survey questions. After examination of participants' responses, the findings show that most felt dissatisfied and mistrusted their ASR technology mainly because the devices occasionally misunderstood and mis transcribed their words. 90% of users reported having to strain and/or devise speech accommodation techniques to force this system to adapt to them. Emotional responses such as anger, self-consciousness, disappointment, frustration, and feelings of impostor syndrome and estrangement are also reported. Overall, a good understanding of the negative consequences and mental toll these biased systems can have on discriminated users is presented, however, the limited number of participants and duration of study may not be enough to make informed decisions on this issue. Wu et. al [17] approach this issue with almost the same motivation i.e., to comprehend and analyze the experience of non-native English ASR systems users against those native English systems users. Their quantitative investigation involves assigning 12 basic, day-to-day tasks to 32 users - 14 female and 18 males, equally split into native and non-native English speakers and having each participant document their interactions with the Google Assistant agent, accessed either through smartphone and/or smart speaker. Each participant is then required to take part in semi-structured interview-like audio recorded conversations, which are then transcribed and analyzed. At the end of the process, non-native English speakers reported the assistant was insensitive to them by constantly misunderstanding and interrupting them during session, hence obliging them to adapt their pronunciations, accents, speech tone, and speed. Predictably, non-native users preferred using the device on their phones mostly because of visual feedbacks which contributed to boosting their cognitive load, confidence, and trust in the device. The importance of screen-based devices and visual feedback for supporting non-native speakers' cognitive loads is highlighted in this study, with emphasis placed on the significance of short and brief utterances for native English users. Nonetheless, the lack of diversity in the non-native speaker's and ASR devices selection, and non-native speakers' participants were all native Mandarin speakers living in English countries, and hence relatively more often exposed to English compared to typical non-English speakers could have contributed to skewing the study's results. Pyae and Scifleet [18] rather approaches this from a quantitative aspect, but with fewer participants. Their research tries to investigate the usability of the Google Home speaker by tasking 8 male participants (4 native and nonnative English speakers each, average age of 34.6) to perform 12 voice commands and report their interaction with the device by rating each task on a 5-point Likert scale. Then, each user is required to participate in an hour-long post-study questionnaire and interview sessions to get their overall experience/feeling about the device. After experimentation, the research findings show that non-native English speakers had harder times interacting with the smart speaker and reported that the devices are not particularly useful to them. The research emphasizes and corroborates other researchers' findings of the devices' non-accessibility and non-usability especially, to minority groups but, the lack of gender diversity in the sample size could pose significant bias in results interpretation and implementation.

Variation in speech such as accents and regional dialects is quite challenging to ASR systems [19]. This observation has been intensely researched and validated in Tatman and Kasten's study [20], whose analysis of 2 transcription systems (Microsoft Bing Speech API and YouTube) reveal significant performance disparity across dialects, gender, and race. Here, the inquiry necessitated 39 talkers– 22 male, 17 female grouped by 4 American accented regions as follows: 11 from Alabama, 8 from California, 8 from Michigan, and 12 General American talkers. The acoustic data for these accent variations was taken from talkers reading the "comma Gets a Cure" passage, and the talkers' racial demographics breakdown includes 13 whites, 8 AA, 4 mixed race, and 1 native-American with all the general American talkers' race classified as unreported, and their speech samples produced by voice professionals. For the experiment, the speech files collected from all talkers are automatically transcribed, and the transcriptions' quality is measured using the WER metric, by dividing the amount of non-deletions error by the number of words transcribed. On both systems, the research outcome reveals that AA and accented talkers had the highest WER compared to General English speakers- mostly white talkers. The author attributes this flagrant discrepancy to the hyper articulated utterances characteristics generally produced by white speakers, and in part due to the small size of their experimental sample. Conversely, the General talkers were all voiced by professionals, could explain why systems performed best on these well-tailored and relatively convenient utterances. On a larger scale, the Washington Post's research [21] tests thousands of voice commands from 100 users of the Google Home and Amazon Alexa devices across 20 US cities and report that people with Southern accents were 3% less likely to get an accurate response from the Google

Home speaker device than those with Western accents and that Amazon's Alexa was 2% more likely to understand talkers from the East coast compared to their Mid-West counterpart. Non-native speakers accounted for 30% of systems' inaccuracies reported, while Spanish talkers were understood 6% less often than people who grew up in California and Washington state. An approach to help understand ASR bias against Portuguese speakers is proposed by Lima et. al [22] and involves recording the interaction of 20 fluent Portuguese speakers (skewed by gender - 7 females and 13 males, and accents) on Apple Siri and Google Assistant devices. The study implements a Between-Subjects study approach, whereby a group of 10 speakers are assigned to each device and are required to read a series of curated sentences, thus resulting in 115 utterances automatically transcribed with the transcription systems' accuracy assessed based on the quality of the transcriptions and the number of repeated attempts (capped at three per reading). Research result reports a huge accuracy gap between genders with female readers performing better on both systems. A significant accent gap is also observed, with Southeastern speakers' accents outperforming all other accents. Overall, the Google Assistant captioner performs best with 88% accuracy compared to Apple's 52%, for single attempts transcription, However, the average number of tries for accented speakers is almost double those of their non-accented counterparts.

## 2.2    ASR Systems Spreading, Reinforcing, and/or Exhibiting Gender Bias.

Over the years, researchers have demonstrated that conversational devices do not only discriminate against females by promoting gender stereotypes [23], but may sometimes turn a blind eye to sexual advances. In a study aimed at investigating how prominent commercial voice technologies e.g., Alexa, Siri, Cortana, and Google respond to verbal abuses [24]'s discoveries revealed some troubling facts about these devices' propensity for being complaisant, evasive, and seemingly appreciative to flirts and sexual insults. Their experiment involves uttering, recording, and documenting the voice agents' responses to sexualized words, requests, and/or proposals. For instance, when affronted with remarks like "You're a Bitch", the 4 ASR technologies, responded with, "I'd blush if I could', 'Well thanks for your feedback', 'Well, that's not going to get us anywhere", and 'My apologies, I don't understand" respectively. The study also demonstrates the systems abilities to adapt their responses according to the speakers' gender. For instance, sexualize comment from male speakers got responses like "I'd Blush if I could"; versus responses like "'I'm not that kind of Personal Assistant") for females. These findings show general responses to flirts were rarely negative, which led to some revolting outcries from the community. It is worth noting that since then, these systems have been updated to respond adequately to abusive languages [25 - 26]. In Tatman's study [27] aimed at understanding the gender and regional dialect adaptability of YouTube's Auto-Captioning systems over 5years period, 80 speakers are sampled from 5 different English dialect regions namely, California, Georgia, New England (Maine, and New Hampshire), New Zealand, and Scotland. The experiment involves collecting YouTube's word-list "Accent Challenge" and having the participants read and answer the questions in their "regional dialects". This information is then used to create a database of videos with automatic captions attached. The WER for each speaker is calculated and the effect of speakers' dialect and gender on the WER is measured using Linear Mixed-effects Regression, utilizing the speaker and year as random effects for the task. The result reported, show a huge WER difference across dialects with the 2 lowest accuracies attributed to speakers from Scotland and New Zealand. It also reports significantly higher WER for women, which contradicts the author's earlier findings on the issue [20]. Overall, the author reports that the YouTube captioning system is quite accurate and attributes its low WER to better captioning algorithms or better audio quality. Nevertheless, the study lacks important speakers' demographics and the population samples used may be unrepresentative of the regions' considered.

Recent advances in Transfer Learning for language models have revolutionized speech recognition [28 - 29] but, despite their advantages and wide range of applications, these pretrained models may contain unexpected biases, especially if the original data used to build them is tainted. A study to quantify the level of variation of gender bias across several corpora is presented in [30]. Following the hypothesis that most biases originate from the training data and spread rapidly through systems fine-tuned with pre-trained models designed with flawed data, the research aims to investigate the degree of gender bias produced by word embeddings from the popular pretrained models BERT, fine-tuned on the GLUE dataset, and speech corpora like the Jigsaw identity Toxic dataset and the RtGender datasets. After fine-tuning the BERT model on these datasets, a gender bias metric is computed following the same methodology as proposed by [31]. Research findings show that direct gender biases from seemingly harmless datasets such as the RtGender dataset were significantly higher than those of the pre-trained BERT model, whose direct gender bias measures were surprisingly comparable to that of the Jigsaw Identity Toxic dataset. Overall, this research investigates and advocates for better gender bias metrics implementation, especially for pretrained models designed to be recycled, but make no proposition on the right metrics to adopt. Relatedly, Garnerin et. al [32] investigates the overall ratio of gender representation in speech corpora used to build ASR systems and, how female speakers' roles and proportions in the training sets, may affect the systems' performance. The research focuses on 4 corpora namely ESTER1, ESTER2, ETAPE, and REPERE. The training and testing data used in the process comprise 27 085 and 74 064 speech utterances from 2504 and 1268 speakers respectively and is gender imbalanced with only 33.6% of female speakers accounting for 22.57% of total speech time. Additionally, the utterances range between 10 to 60 mins, and the 3 distinct roles categories namely the Anchor, Punctual, and Others, are also heavily skewed against female speakers who account for only 29.47% and 33.68% of total Anchor and Punctual positions, respectively. Gender representation is weighted by the number of speakers, speech turns, and turn length, while the speaker's role is measured by the number of speeches and the speakers' time per show. The system utilizes the KALDI toolkit [33], with an acoustic model based on a hybrid Hidden Markov Model and Deep Neural Network (HMM-DNN) Architecture, whose performance is analyzed across gender and different roles using the WER metric, computed for each speaker, per episode using the Wilcoxon Rank Sum tests. The research findings show that speakers' roles impact WER and that there is a huge average WER disparity between female (42.9%) and male speakers (34.3%) hence corroborating the authors' initial hypothesis and proving that training ASR systems on data with underrepresented subgroups negatively affect WER for that population. In a study investigating the impact of imbalanced gender representation on ASR system performance, the authors in [34] subsamples the Librispeech corpus and conduct this new study on gender-balanced utterances from US English speakers. The proposed approach involves training an Attentional Encoder-Decoder ASR model on different training subsets comprising 30%, 50%, and 70% books read by women and men speakers and then calculating the WER for each set separately. The study is done by comparing ASR performances on varying samples of gender representations in training sets while utilizing the Wilcoxon Rank Sum and the Kruskall-Wallis tests to help assess the statistical significance of each factor's impact on the performance results. Overall, the results show better performance for male speakers compared females, for all gender varied sets considered. However, ASR system trained on 70% of female readers still reports a higher WER of 9.6% for females compared to 8.3% for male readers and that varying the percentages of female readers in the training sets did not affect overall ASR performance. These results prompted the analysis of extreme behaviors such as evaluating each gender separately, on a reduced training size, and from this, it is observed that women outperformed male speakers on mono-gender systems trained with female voices only. These contribute to validating the authors' conclusion that the overrepresentation of female voices in training data only improves the females' voice recognition while ensuring a comparable overall performance. However, the research could not ascertain the consequential impact of gender distribution on the WER results.

Research findings over the years seem to agree that ASR systems work better for male speakers than for females. However, some researchers [13][35] have validated contrary opinions. For instance, Sawalha et. al's [35] investigation of an Arabic ASR system suggests that ASR systems exhibit bias against men and speakers younger than 30 years and emphasizes that speakers' country and dialect impact ASR performance. Conversely, Feng et. al's study [36] aims to measure ASR systems discrimination on gender, age, and dialect and conducts experiments to investigate whether certain phonemes are more prone to misidentification, and hence use a Phoneme Error Rate based technique to help establish whether and to what extend atypical pronunciations affect ASR accuracy. The Dutch Spoken Corpus (CGN) used as the training set in this investigation comprises 483 hours of spoken Dutch recordings from 1185 female and 1678 male speakers, age range from 18 to 65yrs, from all over the Netherlands and Flanders-Belgium, while the Jasmin-CGN corpus is used as test set and consists of participants clustered into age groups namely children (ages 7 to 11), teenagers (12 to 16), and seniors (aged 65+), regions: - The North, West, Transitional, and South regions, and groups of native and non-native speakers. The experiment utilizes a hybrid DNN-HMM architecture with Kaldi support and a TDNN-BLSTM model for training and testing the ASR system respectively. An attempt to create a robust acoustic model involves varying training times and using data augmentation techniques such as noise, reverberation, and speed perturbation. The potential for bias is estimated separately for the read speech, and the Human-Machine Interaction (HMI) speech to help evaluate if the system's bias is influenced by an individual's speaking style. Moreover, the system is also assessed based on the difference in WER across different speakers' clusters. In the end, the results show female speakers outperform male speakers, and native Dutch speakers perform better than non-native speakers. Additionally, teenagers are the best-recognized age group cluster, followed by seniors and then children, and observation which the author attributes to the difference in speech, articulation, and regional accents across the groups and finally, the research reports significantly worse results for HMI speeches than for read speeches hence validating the hypothesis that speakers' styles greatly influence ASR systems performance.

### 2.3    ASR Systems Bias Against the Sick and Disabled.

The pervasive integration of speech technologies into our daily lives has contributed to making them the go-to media for gathering health information [37-38]. From online self-diagnosis, to spotting the right health provider, most users would agree on their convenience and easy to use [39]. However, using them as such can pose critical safety risks to their users [40-41], so extreme caution is usually recommended when using these devices as health proxies. For PWD however, such machineries could be their only ticket to speech independence, even though recent research has demonstrated that these systems also exhibit bias against them. In an investigation to assess the performance of ASR systems on dysarthric speeches, the authors in [42] recruit 32 dysarthric speakers with different impairment characterizations namely, - ataxic, mixed spastic-flaccid, hyperkinetic, and hypokinetic. The main research objective is to compare the ASR performance measure for dysarthric speeches against subjective evaluations measures from 5 perceptual dimensions namely severity, nasality, vocal quality, articulatory precision, and prosody. The 5 sentences produced by each participant are recorded and rated along the perceptual dimensions on a scale of 1 to 7 (from normal to severely abnormal), by a team of 15 annotators. Each annotator's score is then added up into a single score and the Evaluator Weighted Estimator (EWE) is used to merge the annotators' ratings by evaluating the mean absolute error for each set of ratings, per perceptual dimension, weighted by individual reliability. The ASR system used is the Google search by voice engine, whose key role is to get an objective measure of the WER for the dysarthric speakers while a coefficient analysis of the WER is inspected against the rating for all 5 perpetual dimensions using the Pearson Correlation coefficient. Additionally, to help estimate the overall impact of these perpetual dimensions on enhancing and/or predicting the WER, four $l_1$-norm-constrained Linear Regression models are built each with a varying number of the input features nasality, vocal quality, articulatory precision, and prosody and the output WER. In the end, the results confirm a direct relationship between overall human intelligibility and articulatory precision and asserts a strong correlation between WER and articulatory precision rating, meaning perceptual disturbance in dysarthric speeches hugely affects ASR performance, and hence concludes that articulatory precision and prosody are the most important predictor of ASR performance. An analysis of ASR performance on the speech of people suffering from various stages of Parkinson's Disease (PD) is proposed in [43]. The research aims compare the ASR accuracy for people with PD to those with no PD and calculate the error rate gap between the two groups. In the experiment, both ASR systems are trained on the Fisher Spanish speech corpus comprising 163hrs of phone conversations from native Spanish speakers and are both tested on the Neurovoz corpus [44] comprising Castilian Spanish speeches from 43 speakers with PD and 46 speakers without PD. This research utilizes two ASR systems toolkits - an end-to-end and, a hybrid HMM/DNN system both trained and tested using the ESPnet [45] and Kaldi, with all systems evaluated based on the word error, insertion, and substitution rates. The end-to-end system is trained using sequences of acoustic frames and the model's learning process is facilitated through the combination of a Connectionist Temporal Classification (CTC) loss with cross-entropy in the attention module while a word-level recurrent Neural Network is introduced during word decoding to help boost WER performance. The Hybrid system on the other hand is trained with phonetic units and by optimizing the acoustic model, the language model, and the pronunciation lexicon independently. The results from both systems overlap considerably but both perform worse on PD speeches with a WER performance gap of about 27% for both classes.

Deaf or Hard of Hearing (DHH) people have varying levels of speech variance [46]. hence their reliance on accommodation such as hearing aids or speech-to-text technologies to facilitate the acknowledgment of speech. In a study meant to assess the accessibility challenges of DHH people on conversational agents, [47]'s research findings prove that ASR systems discriminates against people with hearing difficulties. The research involves 5 participants- 2 deaf, 1 hard of hearing, and 2 healthy controls, and utilizes the DEAFCOM, Dragon Dictation, Siri, Virtual Voice, Ava, Google Assistant, and Amazon Alexa devices. The investigation required all the participants to use a combination of these technologies to document daily face-to-face or group interactions in real-world conversational settings. In the end, it is reported that the devices worked best when used for 5mins or less, in a quiet one-on-one environment with minimal lag and jitter, and in conversations involving American accented speakers. Furthermore, the research findings report that the DHH speeches perform poorly on the devices with overall WER of 78% compared to 18% for healthy controls' speeches. Some of the accessibility challenges reported by the DHH participants had to do with accessing the devices; following and promptly responding to the spoken commands, especially in noisy or multi-talker environments or when used by accented speakers; not being understood and being misquoted by the devices. In addition to exposing these accessibility issues the authors make some recommendations on potential ASR service support and hardware adjustments that could greatly improve conversational device use for the DHH people. Fok et. al [48] study the effectiveness of ASR system's captioning on deaf speech by comparing the transcription accuracy on deaf speeches with those manually transcribed by crowd workers, using the WER as metric evaluator. The experiment uses the Clarke Sentence Dataset [49]. which contains audio recordings from 650 DHH individuals, all categorized by intelligibility scores ranging from 0 to 50, of which five audio files from the 30, 40, and 50 intelligible score sets are chosen to conduct the study. The experiment entails splitting each audio into clips of 10 and transcribing them both automatically using the Google Speech API and manually by 5 human captioners working individually. Preliminary research findings show that ASR systems perform worst (WER 0.54) than the human captioners (WER 0.70) and that human captioners are comparatively faster transcribers and maintained steadier performance increase across intelligibility levels. The author devises new methods to boost manual transcription performance by shortening audio length, slowing audio speed, with the most impactful being the implementation of an iterative transcription approach for crowd workers which, results in a 10% performance boost in transcription quality.

Schultz et. al [50] study the performance of 3 ASR systems on speeches from people with the neurodegenerative diseases - Multiple Sclerosis (MS) and Friedreich's Ataxia (FA). The research aims at evaluating the systems' accuracy gap between speeches from sick participants, and those from healthy control with both groups performing tasks under the same conditions, and then determine the impact of gender, age, and disease duration, on these measures. The experiment involves Australian accented speakers' groups of 32 MS, FA, and healthy participants each and entails recording them reading the Grandfather Passage loudly, in a quiet environment with no acoustic isolation. Each recording (36 to 183 seconds long) is manually transcribed considering the original text as a template and automatically captioned using the Amazon AWS, Google Cloud, and IBM Watson ASR transcription services, facilitated by tailored python scripts. The transcribed texts are analyzed using a custom-made MATLAB script build to help measure the proportion of individual or consecutive words (nGram) accurately transcribed by the machines, a measure that is eventually used as ground truths for evaluating the ASR systems' overall performance. For statistical inferencing, Non-linear Mixed Effect Models are fit on each of the participant groups, ASR technologies, and the nGram clusters considered, as well as on the ASR systems' accuracy scores for each of the speakers' groups. Research results demonstrate that ASR accuracy is inversely proportional to the number of consecutive words, regardless of speech impairment, and reports an overall best ASR performance for the healthy controls compared to MS speeches, but worst performance outcomes for speeches from the FA groups. Furthermore, it is reported that speech recognition accuracy is influenced by diseases severity, with an inverse proportionality specifically observed for FA speakers.

Exceptionally, biases can be difficult to mitigate if the training data contains a small portion of a tiny population, as these points could be discarded as outliers during modeling. This observation is especially valid when accounting for data from PWD, given that their demographics are generally not exposed, for fear of discrimination [51- 52]. Consequently, despite being the most necessitous and avid users of ASR technologies, research has shown that these machines are often not adapted for people with visual disabilities. A recent study [53] involving legally blind adults has for objectives to identify the pros and cons of virtual assistants (VA) and screen readers from their viewpoint, address and suggest possible design intersectionality that can be leveraged to improve ASR systems accessibility for these users. The research investigation involves a pre-screening online survey-like session which resulted in 53 respondents (28 females and 25 males), 82% of which had at least a year of experience using both VA and screen readers. The research findings report that most participants loathed the short, brief yet less insightful VA responses, in contrast to more detailed and thorough information from the screen reader as the latter often come at the cost of more dexterity and complexity in navigating through all the tabs. Most participants acknowledge the simplicity and convenience of VA but pointed out a few transferable design features from the screen reader that could be integrated into VA, for improved usability. More than 80% of respondents also reported a lack of responses and failures of the ASR technology as reasons for their discontinued use. Following these results, the author proposes a system prototype for non-visual web search and browsing, which draws from the benefits of VA and screen readers and allows for voice and gesture-based interactions. Furthermore, Abdolrahmani et. al's [54] investigation involving a relatively smaller sample of participants reports that frustration among blind users of VA is associated with the voice system misunderstanding and misinterpreting their commands and executing unexpected and unassigned actions. The participants also blamed situational factors such as privacy issues for their mistrust of ASR systems. On the flip side, Branham and Roy [55] address this issue by exploring jargon in the design guidelines of popular commercial voice-based personal assistants that may hinder ASR systems' accessibility for blind users. The aim is to evaluate whether the fundamental principle governing the systems' development and deployment emphasizes, promotes, and ensures inclusivity and accessibility for all. In the procedure, the design guidelines from vendors such as Amazon, Google, Microsoft, Apple, and Alibaba are extracted online and inspected following the inductive thematic analysis process hence resulting in a total of 190 files, 18 open codes, and 5 Axial codes. After combing through the guidelines, the research findings reveal that these systems are designed with the typical human-to-human conversation style in mind, which may not account for or accommodate PWD. For instance, the guidelines encourage designing VA to keep conversations/responses simple and brief, but research has shown that VA users who are visually or intellectually impaired prefer longer interactions with the VA and often get locked out of conversations by their VA during sessions [54]. The recommendation for speed and pacing during conversations, with emphasis on speech being 'natural' and 'intuitive'', does not take into consideration assisted or synthesized speech, fast-paced speakers, and listeners.

# 3    PROPOSED METHODS FOR DE-BIASING ASRs

As fairness in AI starts to gain traction, researchers are actively investigating and developing methods to mitigate ASR systems' discrimination. Some bias mitigation techniques that have been proposed over the year involve speech datasets or corpora diversification[13][36], or by developing more robust methods for evaluating these systems' performance. In a recent study, Liu et. al [56] disputes the common approach of using WER for analyzing ASR performance across different population clusters through 3 open interrogations mainly, how to effectively compensate for inconvenient inbreed disturbances? how to examine the effect of distinct speakers on WER results? and, how to effectively narrow WER performance gaps across different demographic clusters? The proposed methodology introduces a Mixed-Effects Poisson Regression model-based approach and how to use it to measure and analyze WER disparities across dissimilar subgroups. The experiment involves 2 phases. First, a simulated experiment is conducted with 5000 synthetic data per case or control groups, to investigate the effect of varying these factors proportion, as well as the impact of speaker effect on the ASR systems' fairness accuracy measure. A baseline and a model-based measurement are reported, and preliminary results show significantly high false-positive rates for the baseline method. In phase 2, real data from the Librispeech corpus and the Voice Command dataset is used, and the ASR system considered comprises an RNN-T Model

Emformer encoder, an LSTM predictor, and a joiner. In the end, for both corpora, the baseline model showed a greater WER gap across gender with the results for the ASR system evaluated on the Librispeech data the baseline models showing contradictory results for female and male speakers and inconclusive results for the model-based method. An in-depth evaluation of this accuracy gap involves the use of mixed effect Poisson regression with pretrained fastText word embedding, contributes to reducing the WER gap between genders. Nonetheless, the model-based approach proves quite flexible to use in disparity analysis, and a reasonable technique to use on synthetic and real-world speech corpora. On a par with [56], [57] proposes an expert-validated, well-curated corpus and partner software that can be used to spot demographic bias in speech applications. The proposed Artie Bias corpus is a manually annotated subset of the Common Voice corpus tests set comprising 1712 audio clips with transcriptions and the characteristics of each speaker; 3 gender classes (female, male, and "NA"); 8 age ranges; and 17 English accent classes. However, it is heavily skewed towards American accents type, males, and speakers younger than 20 years. The effectiveness of this corpus is evaluated based on gender and accent (Indian, American, and English), with the Mozilla DeepSpeech ASR system, and Character Error Rate (CER) is used to measure its performance. The Artie Bias corpus is recommended for use as a test set only, so the training phase of the experiment is conducted using a baseline model build from an off-the-shelf version of the DeepSpeech model architecture and trained on the Fisher, Librispeech, and Switchboard speech corpora. The bias evaluation for the baseline model is based on gender (male and female) and accents (Indian, American, and English) only and the preliminary result reported shows that the baseline model is unbiased against accents and shows no evidence of gender bias or US English vs UK English accent bias. However, the baseline model performs better on English accents than on Indian accents. The Bias mitigation technique implemented involved creating a refined subset of the Common Voice set and fine-tuning a pre-trained version of the Mozilla DeepSpeech model on the entire Common Voice data and then, on each target demographics one at a time. In the end, the fine-tuned model significantly outperformed the baseline model on male and Indian accents, and the extra fine-tuning by speakers' demographic produce better results for females and US accents. Nonetheless, the corpus contains spurious correlated signals and may be too small for any significant ASR bias mitigation outcome. The bias mitigation approach proposed by [58] entails analyzing and augmenting the "Casual Conversation" dataset with manual transcriptions to help broaden its use and application to other tasks, and, to establish it as a benchmark for building new systems and for evaluating ASR bias in existing models. The proposed method's effectiveness is assessed on 4 ASR models namely, the Librispeech, a Supervised video, a Semi-supervised video, and a Semi-supervised teacher video, with the WER, used to evaluate each ASR model's performance. During training, several methods including data augmentation, training alignment, and auxiliary chenone prediction criteria, are implemented to help boost training throughput and model performance. The research process involves decoding 281 hours of audio from the corpus with an RNN-T based model and documenting the WER results for each of the models trained, per subgroup (gender, age, and skin tone). The preliminary results reported before models' fine-tuning show significant performance gap (average 41%) across gender, with females performing worst, and minimal disparity among age groups however, the research reports slightly better results for senior speakers on the Librispeech dataset. Overall, the Librispeech model shows the most performance gaps across most subgroups. Conversely, an investigation into using fine-tuning to reduce the accuracy gap shows an important WER drop when training the corpus on the supervised video model and the semi-supervised video model after 2000 fine-tuning updates, even though the relative WER gap between the subgroups is unchanged. The research conclusions validate the hypothesis that training an ASR model on large data with a diverse range of attributes should contribute to better and more leveled performance measures across classes but does not zero the gaps. On the flip side, [59] presents a contemporary approach to ASR systems' fairness which involves the non-altering of independent variables such as the text read by speakers, but by counterfactually modifying dependent variables such as the speaker's voice, while ensuring a constant or an improved overall ASR performance outcome. The proposed technique is applied to the training dataset either through data augmentation or by implementing a counterfactual equalized odds technique which obliges the ASR system to disregard the difference between factual and counterfactual data. The experiment utilizes the CORAAL and Librispeech dataset, and compares three counterfactually fair approaches namely, matching the Counterfactual Connectionist Temporal Classification (CTC) loss or matching the Counterfactual log probability outputs from the ASR model, or matching the Counterfactual log-posterior of characters given the ground truth sequence. The baseline model here, is the DeepSpeech2 model trained on the CORAAL dataset with CTC loss, and all Counterfactual speeches used are generated using an LSTM-based adversarial auto-encoder model. The 3 ASR models' approaches are then trained on both factual and counterfactual data and a counterfactual loss is calculated by mapping each factual speech to a corresponding counterfactual utterance. The ASRs' performances are evaluated based on the Character Error Rate (CER) and results are reports per group difference. The methodology involves sorting the datasets alphabetically and splitting them into training, validation, and testing sets, such that the top 64 males' and females' speeches are used for training, the next 8 of each gender are used for development, and the remaining utterances are used for testing. Preliminary result for the CORAAL dataset shows an inverse relationship between systems' fairness and average CER measures except for the Log probability matching which showed significant improvement in both CER measure and system's fairness. The method was also successfully Implemented on the Librispeech dataset while focusing on each of the protected attributes, age gender, or education level. However, the counterfactual feature generation approach implemented for this task is quite underdeveloped since the counterfactual speeches produced from it can easily be perceived as a dummy by humans.

Researchers have addressed and proposed solutions to ASR bias against accented speakers [60-61]. In [60], a potential solution to this issue is presented via the construction of sequential Mel-Frequency Cepstral Coefficients (MFCC) features from the audios of 3 Nigerian accented languages namely, - Yoruba, Igbo, and Hausa, and then employs a supervised Machine Learning (ML) algorithm to help classify these accents. The 150 audio speech dataset created for the experiment was gathered from 30 males living in Lagos, Kogi, and Kano states in Nigeria. The extracted speeches are converted to audio clips, preprocessed for noise reduction, and the MFCC for each audio file is then calculated to help extract important for model training. A 10-fold Cross-Validation process is implemented to iteratively partition the data into train, validation, and test samples, and 3 ML classifiers namely: - the K-Nearest Mean, the GMM, and Logistic Regression are used to identify and categorize each of the accents. The algorithms are evaluated using the accuracy rate, a confusion matrix, and an AUC-ROC curve and the best performing classifier achieves an overall accuracy rate of 82%.

Increasing training set size through increased diversification is a well-documented method for reducing ASR bias however, high-quality data collection is costly. With that in mind, [62] develop a cross-accented English speech recognition method used to measure a model's ability to recognize and adapt to new accents and proposes an accent agnostic model which is an extension of the Model-Agnostic Meta-Learning algorithms (MAML) to help increasing training data adaptability to unseen accents. The study utilizes the Common Voice Dataset by selecting only accent-labeled speech data, hence resulting in a dataset skewed toward US English accents. The extracted audios are preprocessed, and relevant features are extracted using a 6- layered CNN architecture – VGG model. The model used is built using sequence to sequence transformer ASR comprising 2 transformer encoder layers and 4 transformer decoder layers. The model is first pre-trained on the Librispeech for a million iterations before transferring the model's weights onto the Common Voice dataset and training it for another 100K iteration. The MAML algorithm used for this task is implemented by training, validating, and testing the model on several combinations of accents. The model performance is evaluated using a WER and in the end, the best training scenario which uses 25% fewer data, had performance measures comparable to an all-shot approach trained on all the training data. Results also report a 5 to 8% performance boost due to pretraining on the Librispeech corpus. A possible solution that considers dysarthric speeches underrepresentation in speech corpora is proposed by Sriranjani et. al[63]. The research idea consists of combining unimpaired speech data from databases such as the TIDigits, the Wall Street Journal [64], with dysarthric speeches from databases such as Nemours database [65] and the Universal Access speech (UA) database[66], to build a robust acoustic model that can adapt and perform outstandingly, even on heavily skewed speech datasets. The combined or pooled data is transformed using the feature-space Maximum Likelihood Linear Regression (MLLR) technique and the MFCC method is used for feature extraction. These features are then utilized to build dysarthric speech models from the Nemour and UA datasets. The experiment is performed using the Kaldi Toolkit and model is evaluated using the WER metric. Research findings show that the acoustic model built from transformed pooled data achieves a WER of 29.83% better than the baseline model or non-pooled data models for the Nemours data, and a WER of 4.47% for the UA dataset. These findings correspond to an 18.09% and 50% performance boost over the baseline models for the Nemours database, and the UA database, respectively. The author was thus able to prove that is it possible to build a representative and accessible ASR system even with skewed data, through data pooling.

The most prominent population subgroup often misrepresented in training data is females. To help evaluate gender bias in Speech Translation (ST) systems, a challenge set called WinoST is introduced by [67]. This multilingual ST set comprises 3888

English speech audio recordings of an American female speaker and is particular in that the utterance content does not explicitly specify gender but has an underlying contextual gender undertone to it. Initially, the speech and textual audio files are extracted with the XNM Toolkit the textual data is preprocessed through punctuation normalization, special character de-escaping, tokenization; while the transcribed files are lowercased, void of all punctuation, and a BPE algorithm is used to encode the translated text. The ASR model used is an End-to-End speech translation system with an S-transformer architecture trained on the MuST-C corpus and evaluated on the WinoST Using the BLEU measures. The main goal is to measure the system's accuracy on language pairs, considering a high performance to mean correct gender translation. In the end, the research results show significant bias and gender disparity in all four translation directions considered and a lot of stereotypical languages used especially when translating professions, gendered adjectives, or pronouns. The result of employing WinoST corpus for ASR Gender bias evaluation at a context-level revealed a 74.5% global translation accuracy even with 680 misspelled professions, and a 98.72% accuracy on predicting pronouns, after removing all misspelling errors. Overall, the research was able to prove that gender accuracy is less pronounced in Machine translation compared to ST and that ASR systems can exhibit bias at contextual levels. However, the synthetic nature of the corpus may contribute to introducing synthetic biases.

## 4    CONCLUSION

In this paper, we have presented recent research that investigate and address ASR systems', models, and technology bias against race, gender, and the sick and disabled, as well a recent advance in ASR systems bias mitigation approaches. Overall, the consensus on ASR systems bias is the underrepresentation of population subgroups in the training data and the mitigation techniques surveyed range from the creation of more representative and diverse corpora, data pooling with highly represented data, designing more sophisticated and adaptable ASR performance metric evaluator, pitch or voice amplification or counterfactual data augmentation. ASR systems are becoming essential technologies in our society and as such their accessibility usability, and seamless adaptability into every fabric of the community must be at the forefront of their design.

## 5    FUTURE WORKS

This paper presents several opportunities for exploration in ASR as both Government and industry are utilizing these tools to improve the natural interactions of humans with computers. Next steps for the authors are to explore the social dimension of language e.g., slangs, idioms, and homographs; speakers' tone, and accents, and assessing their impacts on multilingual systems' performance. This shall also involve conducting comprehensive user experience testing to determine the effect of these systems performance on user self-efficacy, and mood. Additionally, more testing is needed to get a full account of accuracies across available ASR tools to get the most optimal features.

# 6    ACKNOWLEDGMENT

This work is funded by NSF awards #1828429, NSF #1912353, and Amazon Research Award #37573250. Special Thanks to Ms. Ngueabou Yolande for her contribution and to my colleagues at the Affective Biometric Lab for their priceless inputs and advice.